\documentclass[sigconf]{acmart} 

\AtBeginDocument{%
  }

\usepackage{multirow} 
\usepackage{enumitem} 
\usepackage{float}  
\usepackage{pifont} 
\usepackage{subfigure} 
\usepackage{url} 
\usepackage{booktabs} 
\usepackage{tabularx} 

\copyrightyear{2026}
\acmYear{2026}
\setcopyright{cc}
\setcctype{by}
\acmConference[KDD '26]{Proceedings of the 32nd ACM SIGKDD Conference on Knowledge Discovery and Data Mining V.2}{August 09--13, 2026}{Jeju Island, Republic of Korea}
\acmBooktitle{Proceedings of the 32nd ACM SIGKDD Conference on Knowledge Discovery and Data Mining V.2 (KDD '26), August 09--13, 2026, Jeju Island, Republic of Korea}
\acmDOI{10.1145/3770855.3817584}
\acmISBN{979-8-4007-2259-2/2026/08}
\begin{document}

\title{{UrbanWell: Benchmarking Multimodal Large Language Models for Spatio-Temporal Urban Wellbeing Analytics}}


\author{Yanxin Xi}
\affiliation{%
  \institution{Department of Computer Science, University of Helsinki}
  \city{Helsinki}
  \country{Finland}}
\email{yanxin.xi@helsinki.fi}

\author{Xiang Su}
\authornote{Corresponding authors.}
\affiliation{%
  \institution{Department of Computer Science, Department of Agricultural Sciences, University of Helsinki}
  \city{Helsinki}
  \country{Finland}}
\email{xiang.su@helsinki.fi}

\author{Jie Feng}
\authornotemark[1]
\affiliation{%
  \institution{Zhongguancun Academy}
  \city{Beijing}
  \country{China}}
\email{fengjie@bza.edu.cn}

\author{Yu Liu}
\affiliation{%
  \institution{Institute of Biomedical Engineering, Department of Engineering Science, University of Oxford}
  \city{Oxford}
  \country{United Kingdom}}
\email{yu.liu@eng.ox.ac.uk}

\author{Sasu Tarkoma}
\affiliation{%
  \institution{Department of Computer Science, University of Helsinki}
  \city{Helsinki}
  \country{Finland}
  }
\email{sasu.tarkoma@helsinki.fi}

\author{Pan Hui}
\authornotemark[1]
\affiliation{%
  \institution{Computational Media and Arts Thrust, Hong Kong University of Science and Technology (Guangzhou)}
  \city{Guangzhou}
  \country{China}
}

\affiliation{%
  \institution{Department of Computer Science, University of Helsinki}
  \city{Helsinki}
  \country{Finland}
}
\email{panhui@ust.hk}

\renewcommand{\shortauthors}{Yanxin Xi et al.}

\begin{abstract}

Understanding urban wellbeing from multimodal data requires integrating heterogeneous spatial and temporal signals, posing significant challenges for current multimodal large language models (MLLMs). We introduce UrbanWell, a large-scale benchmark designed to systematically evaluate the spatio-temporal reasoning capabilities of MLLMs for urban wellbeing analytics through joint modeling of satellite and street view imagery. UrbanWell spans 38 cities across multiple years and includes diverse indicators covering (1) environmental conditions (CO$_2$, NO$_2$, PM${2.5}$, and Normalized Difference Vegetation Index), (2) spatial accessibility (minimum distance to supermarkets and restaurants), (3) urban form (road length, road density, and land use), (4) urban vitality (population, economic activity diversity, and land use diversity), and (5) subjective perception attributes (e.g., safety, beauty, liveliness, wealth, and quietness). All indicators are aligned at grid level to enable standardized evaluation. Beyond static prediction, UrbanWell defines temporal reasoning tasks, including future value forecasting from historical observations and temporal trend classification. We benchmark 15 state-of-the-art representative MLLMs in a zero-shot setting, providing a comprehensive comparative evaluation across spatial and temporal dimensions. Experimental results indicate that while MLLMs capture salient spatial and perceptual cues, their performance varies substantially across heterogeneous urban indicators spanning environment and subjective perception. UrbanWell serves as a unified benchmark for evaluating multimodal spatial and temporal reasoning in urban wellbeing analytics, offering a standardized testbed for systematic assessment and future research on multimodal urban intelligence. Our codes and datasets are accessible via \url{https://github.com/axin1301/UrbanWell-Benchmark}.
\end{abstract}








\maketitle

\section{Introduction}

Urban areas accommodate over 57\% of the global population \cite{WorldBank_UrbanPop2024}, making cities a major source of large-scale heterogeneous data that reflect environmental, spatial, and socioeconomic dynamics. Understanding multidimensional urban wellbeing requires the integration of diverse signals across space and time, posing substantial challenges for data-driven modeling. Urban wellbeing analytics aims to systematically quantify environmental \cite{andrei2024exploring}, spatial \cite{xi2024pixels}, socioeconomic \cite{fan2023urban}, and perceptual indicators \cite{zhang2018measuring} across neighborhoods, providing a computational perspective on how urban systems function and evolve.

\begin{figure*}[htb]
  \centering
  \includegraphics[width=\linewidth]{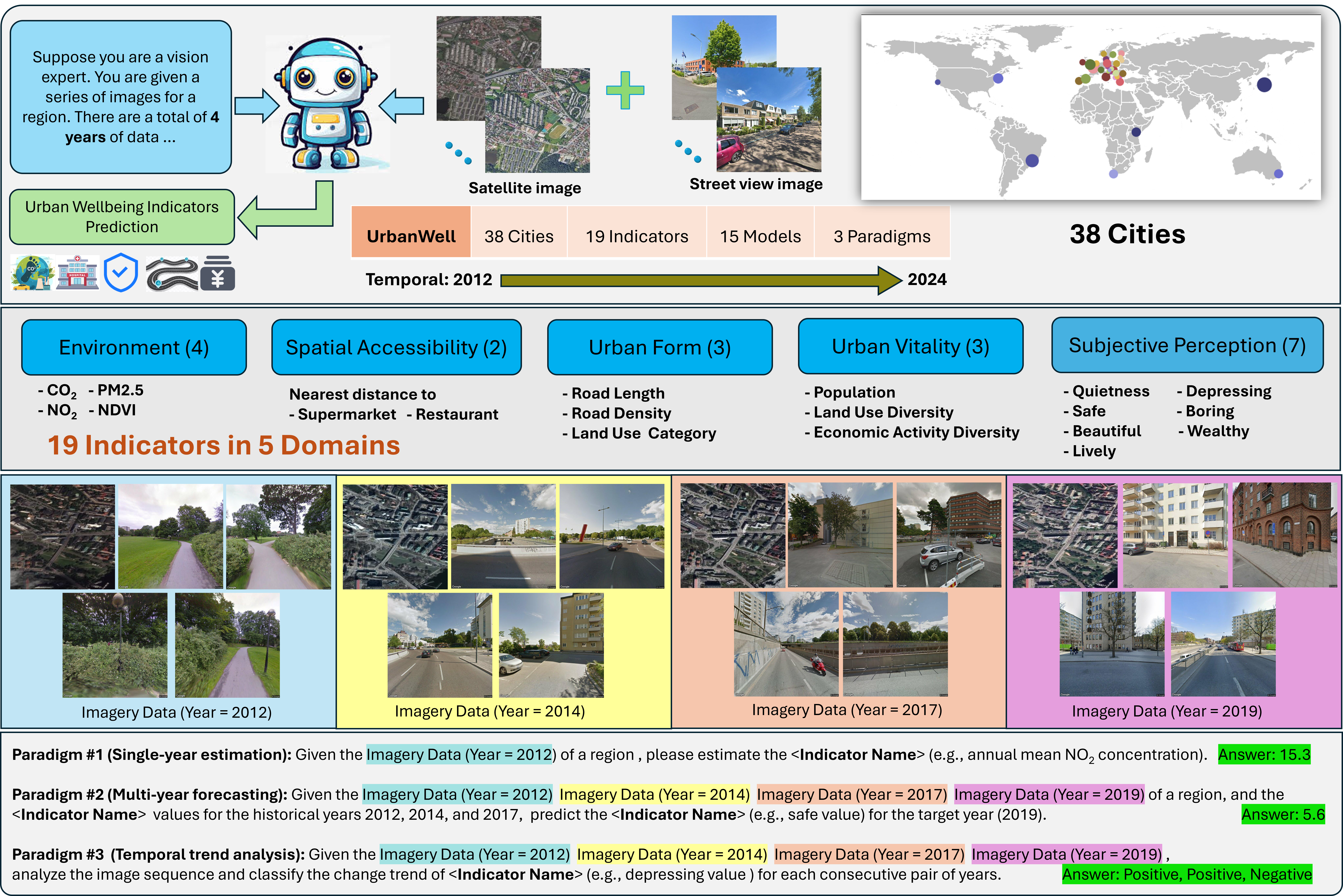}
  \caption{Overview of UrbanWell. UrbanWell integrates satellite and street view imagery with 19 ground-truth urban wellbeing indicators across 38 cities spanning 2012--2024, defining three task paradigms for evaluating the urban intelligence capabilities of 15 MLLMs.}
  \label{framework}
\vspace{-5px}
\end{figure*}

Recent advances in computer vision and urban sensing have shown that deep learning models can infer urban indicators from satellite and street view imagery \cite{li2022predicting,xi2022beyond,liu2023knowledge,fan2023urban}. These approaches estimate attributes such as land use \cite{albert2017using}, inequality \cite{zhang2025perceiving}, and socioeconomic conditions \cite{wang2020urban2vec,sun2025flexireg,jin2025urban,law2019take}. However, existing studies typically focus on isolated tasks, region-specific datasets, or single-year prediction settings, limiting systematic comparison across methods and restricting reproducibility. Moreover, temporal dynamics are rarely incorporated, despite being essential for modeling long-term urban change. The absence of standardized, multimodal benchmarks makes it difficult to comprehensively and quantitatively evaluate model capabilities in spatio-temporal urban reasoning and to assess generalization across cities.


The recent emergence of Multimodal Large Language Models (MLLMs) has introduced a new paradigm for integrating visual and textual information \cite{OpenAIGPT5SystemCard2025, google2025gemini2, team2025gemma}. By jointly modeling images and language, MLLMs enable flexible multimodal reasoning and have shown strong performance across diverse perception and understanding tasks \cite{manvi2023geollm, manvi2024large}. 
However, the capabilities of MLLMs in urban analytics remain underexplored. Existing studies typically evaluate models on limited tasks or single-modality datasets, without systematically assessing cross-city generalization, multi-indicator prediction, or temporal reasoning. Moreover, current urban visual benchmarks are constrained in modality coverage and temporal depth. As summarized in Table~\ref{table-comp}, most of the datasets lack integration of environmental and air quality measures, perceptual attributes, and longitudinal observations, limiting a comprehensive evaluation across spatial and temporal dimensions. These limitations highlight the need for a standardized benchmark that enables systematic assessment of MLLMs for urban wellbeing analytics.

\begin{table}[htb]
  \caption{Comparison with existing urban visual dataset. `SAT': satellite imagery, `STV': street view imagery, `TMP': temporal, `EAQ': environment air quality, `PER': perception, `SC': spatial coverage.}
  \label{table-comp}
  \begin{tabular}{c|cccccc}
    \toprule
    dataset& SAT& STV& TMP& EAQ& PER& SC (\#cities)\\
    \midrule
    CityBench \cite{feng2025citybench} & \textcolor{green}{\ding{51}} &\textcolor{green}{\ding{51}}&& & & 13\\
    CityLens \cite{liu2025citylens}&\textcolor{green}{\ding{51}}&\textcolor{green}{\ding{51}}& & & &17 \\
    UBench \cite{feng2025urbanllava}&\textcolor{green}{\ding{51}}&\textcolor{green}{\ding{51}}& & & &3 \\
    UrBench \cite{zhou2025urbench}&\textcolor{green}{\ding{51}}&\textcolor{green}{\ding{51}}& & & &- \\
    UrbanFeel \cite{he2025urbanfeel}& &\textcolor{green}{\ding{51}}&\textcolor{green}{\ding{51}}&&\textcolor{green}{\ding{51}}&11  \\
    \textbf{UrbanWell}& \textcolor{green}{\ding{51}}& \textcolor{green}{\ding{51}}& \textcolor{green}{\ding{51}}& \textcolor{green}{\ding{51}}& \textcolor{green}{\ding{51}}&38\\
    \bottomrule
\end{tabular}
\vspace{-5px}
\end{table}

To address these limitations, we introduce UrbanWell, a large-scale multimodal benchmark for evaluating the spatio-temporal reasoning capabilities of MLLMs in urban wellbeing analytics. As illustrated in Figure~\ref{framework}, UrbanWell integrates environmental, spatial accessibility, urban form, urban vitality, and subjective perception indicators across 38 cities over multiple years. The dataset aligns heterogeneous signals at grid level and provides a standardized evaluation framework. 
Beyond conventional single-year prediction tasks adopted in prior benchmarks~\cite{feng2025citybench,liu2025citylens,zhou2025urbench,feng2025urbanllava}, UrbanWell defines two additional temporal reasoning tasks, including (1) forecasting future indicator values from historical observations and (2) classifying temporal trends. These tasks explicitly probe the ability of models to perform longitudinal reasoning from multimodal inputs, extending evaluation beyond static estimation settings. Through systematic benchmarking of 15 state-of-the-art MLLMs under a unified protocol, UrbanWell provides a comprehensive assessment of their capacity to model multidimensional spatio-temporal urban signals. UrbanWell is constructed from publicly accessible data sources and released with standardized annotations and task definitions to support comparison across models. Our contributions are threefold:

\begin{itemize}[leftmargin=1.2em, itemsep=0pt, topsep=0pt]

\item We present UrbanWell, a multimodal benchmark covering 38 cities spanning 2012--2024 and integrating 19 indicators across five domains. It provides a standardized evaluation framework for spatio-temporal understanding of urban wellbeing.

\item We formalize single-year estimation, multi-year forecasting, and temporal trend classification tasks to systematically evaluate MLLMs’ spatio-temporal reasoning ability, and benchmark 15 state-of-the-art MLLMs under a unified zero-shot setting.
\item We provide a comprehensive empirical analysis of modality contributions and task design, identifying strengths and limitations of current MLLMs in modeling heterogeneous urban indicators.

\end{itemize}

\section{Dataset Construction}

UrbanWell is designed as a standardized benchmark to evaluate the spatio-temporal reasoning capabilities of MLLMs for multimodal urban analytics. As illustrated in Figure~\ref{framework}, the benchmark integrates 19 heterogeneous indicators across five domains, spanning 38 cities over the period 2012–2024. All indicators are spatially aligned at grid level and temporally synchronized to enable controlled evaluation across static estimation, temporal forecasting, and trend classification settings. The following sections describe the data construction process in detail.



\begin{figure*}[htb]
  \centering
  \includegraphics[width=\linewidth]{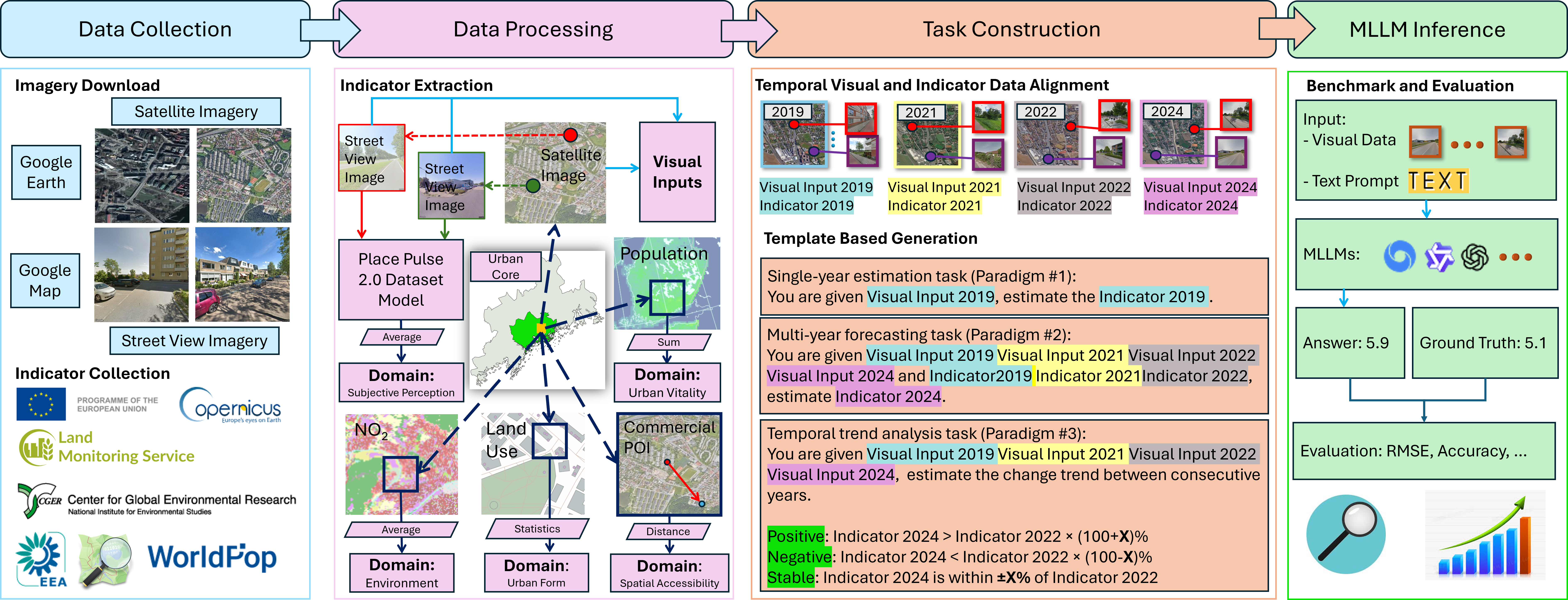}
  \caption{Overview of the UrbanWell benchmark construction pipeline, including data collection, data processing, task construction, and MLLM evaluation.}
  \label{pipeline}
\end{figure*}

\subsection{Data Collection}

The indicator design in UrbanWell follows prior urban analytics literature, with the goal of constructing a heterogeneous and multimodal benchmark rather than focusing on a single predictive task. UrbanWell includes 19 indicators across five domains.

\textbf{Environment} (Carbon Dioxide (CO$_2$), Nitrogen Dioxide (NO$_2$), Particulate Matter 2.5 (PM${2.5}$), Normalized
Difference Vegetation Index (NDVI) \cite{tucker1979red}) provides atmospheric and vegetation-related signals commonly used in environmental urban studies \cite{andrei2024exploring}. \textbf{Spatial accessibility} (minimum distance to restaurants (RTR) and supermarkets (SPM) ) encodes service proximity derived from geospatial data, following accessibility modeling approaches \cite{ganter2022mining}. \textbf{Urban form} (road length (RL), road density (RD), land use category (LUC)) captures structural morphology extracted from OpenStreetMap (OSM) \cite{haklay2008openstreetmap} and land use datasets, consistent with interpretable urban analysis \cite{abitbol2020interpretable}. \textbf{Urban vitality} (population (Popu), land use diversity (LUD), economic activity diversity (EcD)) measures functional intensity and diversity across urban grids \cite{scepanovic2021jane}. \textbf{Subjective perception} ({safe}, {beautiful} (BEA), {lively} (LIV), {boring} (BOR), {depressing} (DEP), and {wealthy} (WEA)~\cite{dubey2016deep}, quietness suitability index (QSI) \cite{eea2014quietareas}) represents perceptual attributes inferred from urban imagery \cite{2024_global_streetscapes, he2025urbanfeel}.


We construct UrbanWell using data from 38 cities, primarily located in Europe and complemented by representative metropolitan areas from Asia, North America, South America, Africa, and Oceania. 
The emphasis on European cities is due to the availability of long-term, standardized, and high-coverage open datasets across environmental, geospatial, and socioeconomic domains, which enable consistent temporal alignment from 2012 to 2024. Cities from other continents are included to introduce distributional diversity and support evaluation under heterogeneous urban conditions.

The list of cities is provided in Appendix~\ref{app_city_names}. Satellite imagery is obtained from Google Earth~\cite{googleearth}, and street view imagery is collected via the Google Street View API~\cite{googlestreetview}. Environmental indicators include CO$_2$ from ODIAC \cite{oda2011very,oda2018open}, NO$_2$ and PM${2.5}$ from the European Environment Agency (EEA)~\cite{eea}, and NDVI from Copernicus~\cite{copernicus}. Urban form and accessibility data include urban core boundaries and land use from Urban Atlas~\cite{urbanatlas2012,urbanatlas2018}, as well as road networks, facility distances, and economic POIs from OSM~\cite{haklay2008openstreetmap}. Population data are obtained from WorldPop \cite{worldpop}. Subjective perception indicators (excluding QSI) are computed using pretrained models applied to large-scale street view imagery, while QSI is obtained from EEA~\cite{eea}.
All data sources are temporally aligned to the same yearly granularity to ensure consistency across tasks.

\subsection{Data Processing}
Figure~\ref{pipeline} illustrates the processing pipeline for harmonizing multi-source data. For each city, we first define the urban core region to ensure spatial consistency across cities. Within this region, satellite imagery, street view imagery, and indicator data are collected for the period 2012--2024. Each satellite image is treated as a basic spatial unit. Given a spatial resolution of approximately 5 meters per pixel and an image size of $256 \times 256$, each spatial unit covers roughly $1.28 \times 1.28$ km, corresponding to an area of approximately $1.6\,\text{km}^2$. This neighborhood-scale granularity balances spatial detail and computational tractability for large-scale benchmarking. All indicators are aggregated to and aligned with this grid to ensure consistent spatial granularity across modalities. For each spatial unit, we extract the geographic boundaries and perform spatial matching with street view imagery using coordinate-based queries. This procedure links aerial and ground-level observations within the same grid cell, enabling multimodal alignment at a unified spatial scale.

To ensure image quality, we apply an automatic filtering strategy to street view imagery. A pretrained semantic segmentation model \cite{zhou2017scene,zhao2017pspnet} estimates the proportion of key outdoor categories (e.g., sky, vegetation, building, road) in each image. Images with low outdoor coverage or excessive indoor content are identified through threshold-based filtering and subsequently verified manually. Indoor or severely occluded scenes are removed to retain representative outdoor samples. The entire processing pipeline is applied uniformly across all cities and years to guarantee consistency. Next, indicators from different domains are aligned to the common spatial grid.

\textbf{Environment.} For each spatial unit and year, raster-based environmental indicators (CO$_2$, NO$_2$, PM${2.5}$, NDVI) are spatially clipped using the grid boundary coordinates, after which the mean pixel value within the clipped region is computed to produce standardized grid-level environmental measurements.

\textbf{Spatial Accessibility.} 
For essential facility accessibility, we consider two POI categories: supermarket and restaurant. For each category, we compute the Euclidean distance between the centroid of each spatial unit and the nearest corresponding POI. The minimum distance to the nearest POI is used as the indicator value.

\textbf{Urban Form.} Annual OSM road network data are clipped to each spatial unit to calculate total RL. RD is computed by normalizing RL by the grid area. LUC within each grid is assigned as the land use indicator for that year.

\textbf{Urban Vitality.} LUD is computed using an entropy-based measure \cite{scepanovic2021jane} over land use categories within each spatial unit, including (i) residential areas, (ii) commercial, industrial, institutional, and governmental areas, and (iii) recreational spaces, parks, and water bodies. EcD is calculated using Shannon entropy~\cite{shannon1948mathematical} over commercial POI categories (e.g., shops, restaurants, hotels) extracted from OSM within each spatial unit. 
The formulations of LUD and EcD are provided in Appendix~\ref{data-processing-appendix}. Popu is obtained by mapping the population to each grid and summing the values within the region to derive total population per spatial unit and year.

\textbf{Subjective Perception.} 
Pretrained models on the Place Pulse 2.0 dataset \cite{2024_global_streetscapes, dubey2016deep} are applied to street view images located in each spatial unit. For each image, the models output perception scores including {safe}, BEA, LIV, BOR, DEP, and WEA. Grid-level perception indicators are computed as the mean score across all matched street view images within the spatial unit. QSI is obtained directly from the source dataset and matched to the corresponding spatial units.

\subsection{Task Construction}
The construction of benchmark tasks from the processed imagery and indicator data is illustrated in Figure~\ref{pipeline}. To systematically evaluate spatial--temporal reasoning in MLLMs, we define three task paradigms, i.e., \textit{Single-year Estimation}, \textit{Multi-year Forecasting}, and \textit{Temporal Trend Analysis}. 

\textbf{Single-year Estimation Task.}  
Given satellite and street view images of a spatial unit from the same year, the model predicts predefined urban indicators for that year. This task evaluates cross-modal spatial reasoning in a temporally static setting.

\textbf{Multi-year Forecasting Task.}  
Given satellite and street view images from multiple consecutive years for a spatial unit, together with historical indicator values, the model predicts the indicator value in the final year of the sequence. This task evaluates temporal reasoning and longitudinal pattern modeling from multimodal observations.

\textbf{Temporal Trend Analysis Task.}  
Given sequential imagery of the same spatial unit across consecutive years, the model classifies whether the indicator exhibits a positive trend, a negative trend, or a stable trend (i.e., no significant change). This task evaluates the model’s ability to detect temporal dynamics rather than estimate absolute values.

Due to computational constraints, we adopt a stratified sampling strategy to construct the benchmark dataset while preserving geographic and temporal coverage. For each indicator, samples are drawn approximately evenly across all 38 cities. Within each city, grid-year instances are sampled from the available years to maintain temporal diversity. This design ensures broad coverage across geographic regions and time periods.


\begin{figure}[htb]
  \centering
  \includegraphics[width=\linewidth]{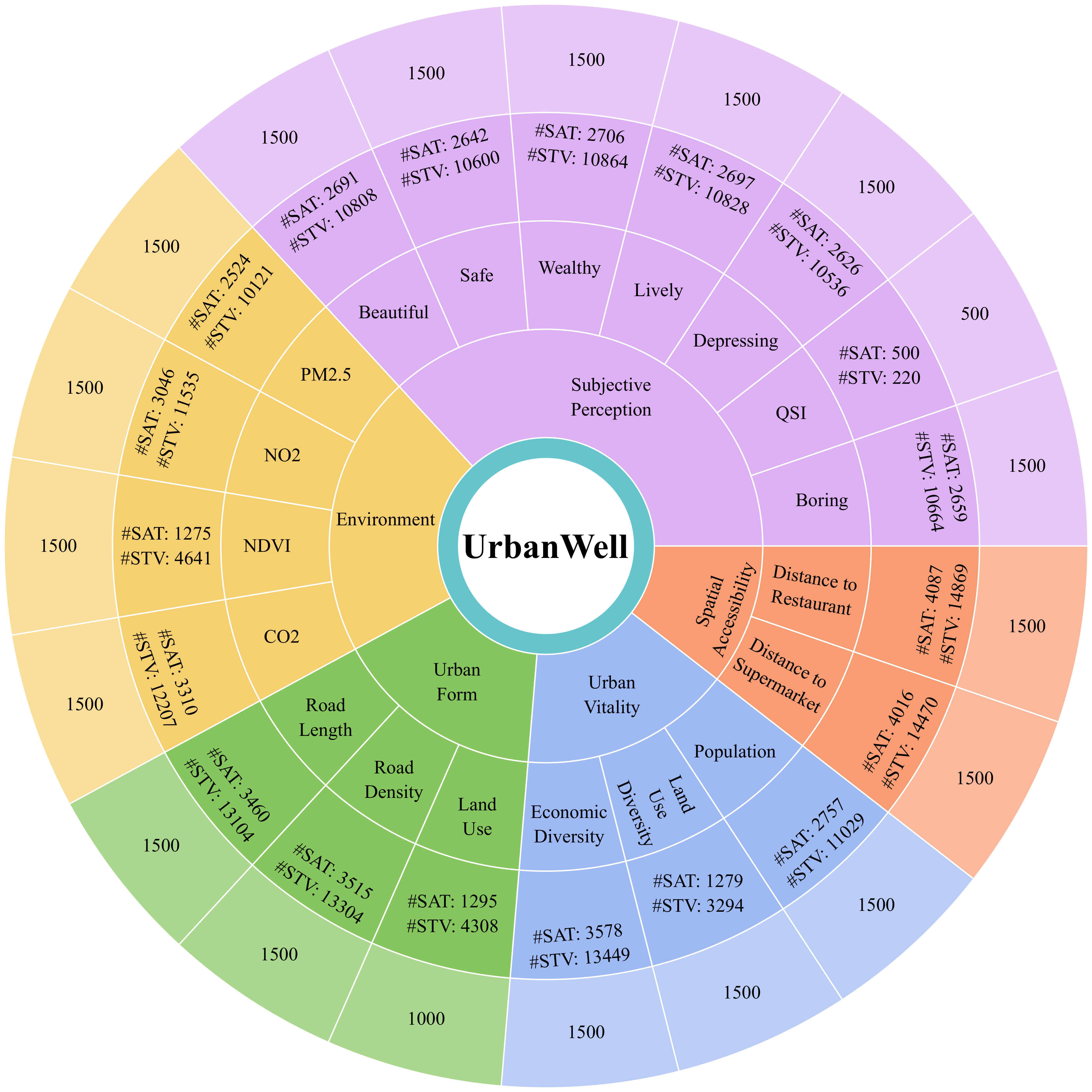}
  \caption{Summary statistics of UrbanWell, showing the numbers of samples, satellite images, and street view images across the 19 urban indicators. Statistics are aggregated over all task paradigms.}
  \label{dimension}
\vspace{-13px}
\end{figure}

\textbf{Dataset Statistics.}
Figure~\ref{dimension} summarizes the number of samples, satellite images, and street view images across the 19 indicators in 5 domains. These statistics provide an overview of benchmark scale and modality coverage across indicators.

\subsection{MLLM Inference}
We evaluate all models under a unified zero-shot setting using task-specific prompt templates. Prompts are constructed according to the three task paradigms and provide the corresponding multimodal inputs, historical observations, and task instructions. For temporal trend analysis, prompts additionally define explicit percentage thresholds for classifying trends as positive, negative, or stable. All prompts follow fixed templates with placeholders for years, image sequences, indicator names, units, and output formats. This standardized schema ensures consistent inputs and outputs across model evaluation. Example prompts are provided in Appendix~\ref{prompt-example}.

Land use prediction is formulated as a multiple-choice classification task. Temporal trend analysis is formulated as a three-class classification problem. All other indicators in the single- and multi-year tasks are treated as regression problems, where models output scalar values. This formulation enables standardized evaluation across heterogeneous urban indicators while maintaining task-specific prediction objectives.

\begin{table*}[htb]
  \caption{Results of single-year estimation tasks. Accuracy is reported for the LUC task, while RMSE is reported for all other indicators. The best result for each indicator is highlighted in bold, and the second-best result is underlined.}
  \label{tablesingleyear}
\centering
\scriptsize
\setlength{\tabcolsep}{3pt}
\resizebox{\textwidth}{!}{%
\begin{tabular}{c|cccc|cc|ccc|ccc|ccccccc}
\hline
 & \multicolumn{4}{c}{\textbf{Environment}} & \multicolumn{2}{c}{\textbf{Accessibility}} & \multicolumn{3}{c}{\textbf{Urban Form}} & \multicolumn{3}{c}{\textbf{Urban Vitality}} & \multicolumn{7}{c}{\textbf{Subjective Perception}} \\
Model & CO$_2$ & NO$_2$ & PM2.5 & NDVI & RTR & SPM & RD & RL & LUC & Popu & LUD & EcD & QSI & Safe & LIV & WEA & BEA & BOR & DEP \\
\hline
Gemma3-4B & 6.2e3 & 19.34 & 15.45 & 0.75 & 6.94 & 6.54 & 17.69 & 16.76 & \underline{0.60} & 1.2e5 & \textbf{0.31} & 1.15 & 4.5e2 & 4.84 & 3.45 & 2.80 & 2.35 & 8.30 & 7.32 \\
Gemma3-12B & 1.5e4 & 18.67 & 3.93 & 0.87 & 7.17 & 7.66 & 16.63 & 17.67 & 0.55 & 7.4e4 & 0.43 & 0.98 & 4.8e2 & \underline{1.83} & \textbf{1.66} & \textbf{1.69} & \underline{2.16} & 4.16 & \underline{4.49} \\
Gemma3-27B & 2.5e3 & 12.75 & \underline{3.68} & 0.86 & 6.27 & 4.27 & 16.19 & \underline{13.82} & 0.59 & \textbf{1.7e4} & 0.47 & \textbf{0.81} & \textbf{3.7e2} & 1.99 & 2.86 & \underline{2.15} & 2.64 & 6.65 & 7.81 \\
\hline

Qwen2.5VL-3B & 1.0e14 & 6.2e2 & 93.21 & 0.80 & 3.6e2 & 77.71 & 2.2e2 & 70.51 & 0.40 & 9.6e6 & 0.52 & 1.82 & 4.3e2 & 7.27 & 6.45 & 5.47 & 6.73 & 9.64 & 9.19 \\
Qwen2.5VL-7B & 1.1e8 & 18.83 & 10.28 & 0.86 & 10.87 & 8.07 & 16.43 & 36.40 & 0.51 & 7.2e4 & 0.53 & 1.48 & 5.7e2 & 3.99 & 4.17 & 4.38 & 3.35 & 7.34 & 7.48 \\
Qwen2.5VL-32B & 5.4e4 & 9.32 & 4.24 & 0.80 & 47.63 & 8.47 & 17.07 & 16.69 & 0.48 & 7.8e4 & 0.41 & 1.42 & 6.2e2 & 3.13 & 4.64 & 5.03 & 4.46 & 8.95 & 7.61 \\
\hline
InternVL3-9B & 3.8e3 & 19.78 & 15.89 & 0.87 & 7.64 & 7.59 & 16.86 & 15.11 & 0.44 & 7.6e4 & 0.45 & 1.41 & 5.8e2 & 3.82 & 5.08 & 4.25 & 3.21 & 7.09 & 9.44 \\
InternVL3.5-14B & 2.4e4 & 16.65 & 5.15 & 0.72 & 7.30 & 6.34 & 15.10 & 15.41 & 0.41 & 2.4e4 & 0.44 & 1.23 & \underline{4.2e2} & \textbf{1.82} & 2.28 & 2.98 & 2.40 & \textbf{3.62} & \textbf{3.90} \\
\hline
Phi-3.5 & \underline{1.4e3} & 29.96 & 19.85 & 0.90 & 4.77 & 6.94 & 15.61 & 27.36 & 0.40 & 1.5e5 & 0.54 & 1.43 & 4.7e2 & 4.72 & 4.43 & 3.86 & 3.48 & 8.72 & 10.89 \\
Phi-4 & 2.8e3 & 69.07 & 63.68 & 0.91 & 4.92 & 5.40 & \underline{14.51} & 1.4e2 & 0.25 & 2.4e7 & 0.54 & 1.02 & 5.4e2 & 9.31 & 7.54 & 8.19 & 8.86 & 10.96 & 11.16 \\
\hline
Mistral-Small-3.1-24B & 6.3e7 & 7.60 & 4.57 & \underline{0.71} & 7.50 & 6.05 & 17.65 & 16.69 & 0.52 & \underline{1.8e4} & 0.42 & 1.30 & 5.6e2 & 2.65 & 4.55 & 5.10 & 3.91 & 4.42 & {6.06} \\
Llama4-Scout & 6.3e4 & 8.78 & {3.92} & \textbf{0.69} & \underline{2.57} & \textbf{2.33} & 16.84 & 17.86 & 0.07 & 7.4e4 & 0.46 & 0.96 & 5.1e2 & 2.62 & 3.54 & 4.64 & 3.98 & 6.21 & 6.41 \\
Nova-lite-v1 & 5.7e6 & 15.29 & 6.66 & 0.78 & 6.40 & 5.48 & 17.07 & 16.99 & 0.49 & 6.9e4 & 0.47 & \underline{0.87} & 6.1e2 & 2.43 & \textbf{2.09} & 2.34 & \textbf{2.06} & \underline{3.86} & 6.95 \\
\hline
Gemini-2.0-Flash & \textbf{2.2e2} & \textbf{5.50} & \textbf{2.98} & \textbf{0.69} & {2.61} & \underline{2.66} & 16.15 & 17.27 & \textbf{0.66} & {2.3e4} & \underline{0.39} & 1.19 & 4.7e2 & 3.67 & 3.60 & 4.29 & 3.50 & 7.18 & 6.74 \\

GPT-5-nano & 1.3e4 & \underline{7.30} & 4.10 & 0.80 & \textbf{2.52} & 3.42 & \textbf{12.79} & \textbf{13.62} & 0.58 & 3.1e4 & 0.57 & 1.21 & 5.4e2 & 2.19 & 5.44 & 3.86 & 2.83 & 5.12 & 7.05 \\
\hline
\end{tabular}
}
\end{table*}

\section{Benchmark and Experiments}

\subsection{Experimental Setup}

\textbf{Model Deployment.} 
We evaluate 15 representative MLLMs, covering both open-source and proprietary systems across different model families and parameter scales. Following \cite{liu2025citylens}, the evaluated models include Gemma3-4B/12B/27B \cite{team2025gemma}, Qwen2.5VL-3B/7B/32B \cite{bai2025qwen2}, InternVL3-9B \cite{zhu2025internvl3}, InternVL3.5-14B \cite{wang2025internvl35}, Phi-3.5-vision \cite{abdin2024phi3technicalreporthighly}, Phi-4-multimodal \cite{abdin2024phi}, Mistral-small-3.1-24B \cite{MistralAIMS312025}, Llama4-Scout \cite{MetaAILlama42025}, Amazon-Nova-Lite \cite{amazon2025nova}, Gemini-2.0-Flash \cite{google2025gemini2}, and GPT-5-nano \cite{OpenAIGPT5SystemCard2025}. All models are evaluated using identical prompt templates and inference configurations. No task-specific hyperparameter tuning is performed.

\textbf{Evaluation Metrics.} 
Evaluation metrics are determined according to task type. LUC is evaluated using classification accuracy. Temporal trend analysis is evaluated using three-class accuracy, with accuracy computed over all predicted trend labels. All remaining indicator estimation tasks are evaluated using Root Mean Squared Error (RMSE). RMSE is reported in the original measurement units, and results are presented separately for each indicator due to differences in scale and semantics across tasks.

\subsection{Performance on Single-year Estimation Task}


\textbf{Single-year estimation remains challenging and exhibits clear indicator stratification.}
Table~\ref{tablesingleyear} reveals substantial instability in quantitative prediction from static multimodal inputs. Environmental indicators are particularly challenging. Although RMSE values are not directly comparable across indicators due to differences in measurement scales (e.g., CO$_2$ is reported in monthly carbon tonnes, whereas NO$_2$ is measured in $\mu$g/m$^3$), CO$_2$ exhibits extreme cross-model variation, with RMSE spanning multiple orders of magnitude. This suggests substantial difficulty in inferring emission-related quantities from visual observations and points to calibration challenges in mapping visual cues to physically meaningful emission scales. NO$_2$ and PM${2.5}$ also exhibit noticeable performance variation across models, indicating that air-quality estimation remains challenging even when environmental signals are partially observable. Urban form metrics (RL and RD) yield relatively consistent performance across models, indicating partial alignment between visible spatial structures and numerical measurements. Popu consistently exhibits large prediction errors, highlighting the limited direct correspondence between imagery and aggregated demographics. In contrast, several urban vitality and perception-related attributes (e.g., economic diversity, BEA, and LIV) exhibit comparatively lower cross-model variance, suggesting that some higher-level urban characteristics can be inferred more consistently from visual cues.


\begin{table*}[htb]
  \caption{Results of multi-year forecasting tasks. Accuracy is reported for the LUC task, while RMSE is reported for all other indicators. For each indicator, the best value across all models is highlighted in bold, and the second-best value is underlined.}
  \label{tablemultitype1}
\centering
\scriptsize
\setlength{\tabcolsep}{3pt}
\resizebox{\textwidth}{!}{%
\begin{tabular}{c|cccc|cc|ccc|ccc|cccccc}
\hline
 & \multicolumn{4}{c}{\textbf{Environment}} & \multicolumn{2}{c}{\textbf{Accessibility}} & \multicolumn{3}{c}{\textbf{Urban Form}} & \multicolumn{3}{c}{\textbf{Urban Vitality}} & \multicolumn{6}{c}{\textbf{Subjective Perception}} \\
Model & CO$_2$ & NO$_2$ & PM2.5 & NDVI & RTR & SPM & RD & RL & LUC & Popu & LUD & EcD & Safe & LIV & WEA & BEA & BOR & DEP \\
\hline
Gemma3-4B & \textbf{11.51} & 3.82 & 2.71 & 0.50 & 0.38 & 0.35 & 5.37 & 5.02 & 0.11 & 8.3e2 & 0.13 & 0.33 & \textbf{1.30} & \textbf{1.23} & 1.25 & 1.49 & 1.04 & 1.25 \\
Gemma3-12B & 13.33 & 2.93 & 2.23 & 0.58 & 0.32 & {0.30} & 4.91 & 5.08 & 0.44 & 2.5e2 & 0.07 & \underline{0.29} & 1.44 & 1.42 & 1.37 & 1.62 & 1.03 & 1.33 \\
Gemma3-27B & 13.99 & 3.06 & 2.24 & 0.39 & 0.32 & \underline{0.27} & 4.46 & 4.80 & 0.61 & \textbf{2.0e2} & 0.07 & \textbf{0.28} & 1.45 & 1.41 & 1.37 & 1.56 & 1.05 & 1.34 \\
\hline
Qwen2.5VL-3B & 27.60 & 3.49 & 2.59 & 0.48 & 0.40 & 0.43 & 5.94 & 5.42 & 0.21 & 3.6e2 & 0.18 & 0.42 & 1.33 & 1.34 & 1.26 & 1.47 & \textbf{0.98} & 1.23 \\
Qwen2.5VL-7B & 14.20 & 3.67 & 2.71 & 0.43 & 0.34 & 0.30 & 4.81 & 4.98 & 0.59 & 3.7e2 & 0.11 & 0.30 & 1.34 & \underline{1.26} & 1.27 & \underline{1.42} & 1.03 & 1.27 \\
Qwen2.5VL-32B & 12.17 & 2.80 & 2.11 & 0.52 & \underline{0.30} & 0.30 & 4.54 & 4.79 & \textbf{0.66} & 2.5e2 & 0.09 & 0.30 & 1.42 & 1.36 & 1.33 & 1.52 & 1.04 & 1.31 \\
\hline
InternVL3-9B & 14.27 & 2.71 & 2.09 & 0.47 & 0.31 & 0.28 & 4.76 & 5.07 & 0.45 & 2.5e2 & 0.10 & 0.32 & 1.48 & 1.49 & 1.45 & 1.65 & 1.16 & {1.41} \\
InternVL3.5-14B & 13.93 & 2.79 & 1.97 & 0.59 & 0.40 & 0.29 & 4.78 & 5.38 & 0.33 & 2.6e2 & 0.11 & \underline{0.29} & 1.46 & 1.46 & 1.43 & 1.59 & 1.09 & 1.35 \\
\hline
Phi-3.5 & 13.12 & 3.50 & 2.73 & 0.60 & 0.33 & 0.33 & 5.00 & 5.28 & 0.18 & 2.8e2 & 0.08 & 0.32 & 1.41 & 1.34 & 1.31 & 1.50 & 1.03 & 1.23 \\
Phi-4 & 14.24 & 3.40 & 2.72 & 0.51 & 0.34 & \textbf{0.24} & 5.33 & 5.24 & 0.10 & 3.0e2 & \textbf{0.05} & 0.30 & \underline{1.32} & 1.30 & \underline{1.23} & 1.49 & 1.04 & 1.26 \\
\hline
Mistral-Small-3.1-24B & 12.66 & \underline{2.48} & \underline{1.96} & 0.60 & \textbf{0.20} & \textbf{0.24} & 5.09 & \underline{4.48} & 0.58 & 4.1e2 & 0.12 & 0.31 & 1.34 & 1.39 & 1.40 & 1.55 & 1.01 & 1.23 \\
Llama4-Scout & \underline{12.10} & \textbf{2.42} & \textbf{1.92} & \underline{0.35} & 0.33 & 0.29 & \textbf{2.56} & \textbf{4.09} & 0.08 & 3.1e2 & 0.21 & \underline{0.29} & 1.33 & 1.30 & \textbf{1.16} & \textbf{1.35} & \underline{0.99} & \textbf{1.14} \\
Nova-lite-v1 & 14.75 & 2.62 & 1.99 & 0.46 & 0.33 & 0.30 & 4.61 & 4.92 & 0.48 & 2.5e2 & \underline{0.06} & 0.30 & 1.38 & 1.33 & 1.32 & 1.46 & 1.06 & 1.27 \\
\hline
Gemini-2.0-Flash & 13.93 & 2.73 & 2.16 & \textbf{0.34} & 0.32 & 0.41 & 4.90 & 5.01 & \underline{0.63} & 2.4e2 & \underline{0.06} & \textbf{0.28} & 1.39 & 1.36 & 1.32 & 1.47 & \underline{0.99} & \underline{1.22} \\
GPT-5-nano & 12.54 & 2.97 & 2.12 & 0.59 & 0.45 & 0.32 & \underline{4.42} & 5.04 & 0.52 & \underline{2.2e2} & 0.23 & 0.36 & 1.46 & 1.38 & 1.44 & 1.54 & 1.08 & 1.33 \\
\hline
\end{tabular}
}
\end{table*}

\textbf{Model performance is shaped by both parameter scale and architectural design.}
Performance differences across MLLMs are substantial. Within several model families (e.g., Gemma and Qwen), larger variants generally achieve lower errors, although improvements are not uniformly observed across all indicators. This suggests that increased model capacity can benefit quantitative estimation. However, cross-family comparisons indicate that architecture and alignment strategy remain equally important. Gemini-2.0-Flash achieves the strongest performance on several environmental indicators and LUC, while GPT-5-nano performs particularly well on urban form indicators, attaining the lowest errors for both RL and RD. Other models exhibit strengths on specific perception and vitality-related indicators. The absence of a universally dominant model suggests that multimodal numerical reasoning remains highly task-dependent rather than being determined solely by model scale.

\subsection{Performance on Multi-year Forecasting Task}

\textbf{Temporal context substantially improves numerical stability.}
Table~\ref{tablemultitype1} presents results on the multi-year forecasting task. Compared to single-year estimation, error magnitudes decrease dramatically and become tightly bounded across models, eliminating the extreme outliers previously observed for CO$_2$ and Popu. This indicates that incorporating temporal context significantly improves calibration for emission-related and demographic quantities. Moreover, performance differences across model families are noticeably reduced, suggesting that temporal context provides a strong signal that partially compensates for architectural differences. 

\textbf{Forecasting difficulty remains indicator-dependent.}
Despite improved stability, substantial differences persist across indicator types. Accessibility indicators and EcD generally achieve more consistent forecasting performance, suggesting relatively regular temporal dynamics. In contrast, Popu remains among the most challenging indicators, while environmental forecasting performance varies considerably across variables. These patterns indicate that indicators influenced by complex external factors, such as demographic change and emissions, remain difficult to predict even when historical multimodal observations are available. By comparison, indicators more directly tied to observable urban structure exhibit greater temporal regularity and are therefore more amenable to forecasting.

\textbf{Temporal context narrows performance differences across models.} Compared with the single-year estimation task, inter-model performance gaps become narrower across most indicators. While larger variants within the same model family sometimes achieve lower forecasting errors, scaling benefits are less consistently observed than in single-year estimation. Notably, forecasting performance becomes noticeably more homogeneous across models, suggesting that temporal context provides a strong predictive signal that narrows performance differences across architectures. As a result, scaling benefits become less pronounced in the forecasting setting.

\begin{table*}[htb]
  \caption{Results of temporal trend analysis tasks. Accuracy is reported for all indicators. For each indicator, the best value across all models is highlighted in bold, and the second-best value is underlined.}
  \label{tablemultitype3}

\centering
\scriptsize
\setlength{\tabcolsep}{3pt}
\resizebox{\textwidth}{!}{%
\begin{tabular}{c|cccc|cc|cc|ccc|cccccc}
\hline
 & \multicolumn{4}{c}{\textbf{Environment}} & \multicolumn{2}{c}{\textbf{Accessibility}} & \multicolumn{2}{c}{\textbf{Urban Form}} & \multicolumn{3}{c}{\textbf{Urban Vitality}} & \multicolumn{6}{c}{\textbf{Subjective Perception}} \\
Model & CO$_2$ & NO$_2$ & PM2.5 & NDVI & RTR & SPM & RD & RL & Popu & LUD & EcD & Safe & LIV & WEA & BEA & BOR & DEP \\
\hline
Gemma3-4B & 0.32 & \underline{0.36} & \textbf{0.41} & \underline{0.53} & \textbf{0.45} & \textbf{0.42} & 0.23 & 0.23 & 0.36 & 0.12 & 0.35 & 0.34 & 0.32 & 0.30 & \underline{0.37} & 0.33 & \textbf{0.37} \\
Gemma3-12B & 0.26 & 0.25 & 0.21 & 0.06 & 0.30 & 0.33 & 0.30 & 0.25 & 0.38 & \underline{0.80}& 0.27 & 0.15 & 0.13 & 0.13 & 0.15 & 0.20 & 0.19 \\
Gemma3-27B & 0.23 & 0.13 & 0.14 & 0.09 & \underline{0.37} & 0.33 & 0.36 & 0.33 & 0.39 & 0.77 & 0.28 & 0.09 & 0.10 & 0.13 & 0.11 & 0.15 & 0.21 \\
\hline
Qwen2.5VL-3B & \underline{0.37} & \textbf{0.38} & \underline{0.35} & 0.44 & 0.36 & \underline{0.35} & 0.32 & 0.30 & 0.36 & 0.17 & 0.34 & \underline{0.37} & \underline{0.36} & \underline{0.35} & 0.36 & \textbf{0.35} & \textbf{0.37} \\
Qwen2.5VL-7B & 0.24 & 0.21 & 0.18 & 0.12 & 0.31 & 0.33 & 0.33 & 0.32 & 0.37 & 0.64 & 0.28 & 0.13 & 0.15 & 0.13 & 0.15 & 0.16 & 0.16 \\
Qwen2.5VL-32B & 0.26 & 0.25 & 0.20 & 0.08 & 0.31 & 0.32 & 0.31 & 0.31 & \underline{0.40} & 0.77 & 0.27 & 0.12 & 0.10 & 0.09 & 0.12 & 0.17 & 0.16 \\
\hline
InternVL3-9B & 0.23 & 0.17 & 0.17 & 0.10 & 0.33 & 0.33 & 0.32 & 0.34 & 0.38 & 0.67 & 0.27 & 0.17 & 0.19 & 0.20 & 0.22 & 0.24 & 0.19 \\
InternVL3.5-14B & 0.30 & 0.28 & 0.31 & 0.30 & 0.31 & 0.32 & 0.38 & 0.39 & 0.39 & 0.63 & 0.34 & 0.31 & 0.30 & 0.30 & 0.32 & 0.30 & 0.29 \\
\hline
Phi-3.5 & 0.23 & 0.23 & 0.20 & 0.27 & 0.32 & 0.31 & 0.33 & 0.35 & 0.37 & 0.68 & 0.30 & 0.25 & 0.30 & 0.26 & 0.25 & 0.27 & 0.28 \\
Phi-4 & 0.30 & 0.27 & 0.27 & \textbf{0.54} & 0.32 & 0.32 & 0.43 & 0.43 & 0.37 & 0.26 & \textbf{0.40} & 0.33 & 0.35 & 0.32 & 0.31 & 0.27 & 0.30 \\
\hline
Mistral-Small-3.1-24B & 0.00 & 0.31 & 0.20 & 0.14 & 0.00 & 0.00 & \textbf{0.56} & \textbf{0.67} & 0.36 & \textbf{0.89} & 0.00 & 0.21 & 0.21 & 0.19 & 0.19 & 0.21 & 0.19 \\
Llama4-Scout & \textbf{0.67} & 0.00 & 0.04 & 0.00 & 0.31 & 0.31 & 0.37 & 0.36 & \textbf{0.52} & 0.75 & 0.36 & 0.21 & 0.19 & 0.16 & 0.20 & 0.21 & 0.16 \\
Nova-lite-v1 & 0.33 & 0.28 & 0.30 & 0.31 & 0.29 & 0.29 & 0.45 & \underline{0.46} & 0.39 & 0.45 & 0.37 & 0.31 & 0.34 & 0.34 & 0.35 & 0.29 & 0.28 \\
\hline
Gemini-2.0-Flash & 0.21 & 0.11 & 0.10 & 0.05 & 0.30 & 0.31 & 0.32 & 0.32 & \underline{0.40} & 0.79 & 0.26 & 0.08 & 0.08 & 0.08 & 0.08 & 0.11 & 0.07 \\
GPT-5-nano & 0.31 & 0.32 & 0.23 & 0.33 & \underline{0.37} & 0.34 & \underline{0.48} & 0.42 & 0.35 & 0.37 & \underline{0.39} & \textbf{0.38} & \textbf{0.38} & \textbf{0.40} & \textbf{0.39} & \underline{0.34} & \underline{0.36} \\
\hline
\end{tabular}
}

\end{table*}

\subsection{Performance on Temporal Trend Analysis}

\textbf{Temporal trend analysis remains challenging.} Table~\ref{tablemultitype3} presents element-wise accuracy for multi-year trend analysis. For many indicators, accuracy remains modest, with a large proportion of results concentrated around 0.2--0.4. Unlike value forecasting, trend analysis requires models to identify the direction of change from image sequences and capture subtle inter-annual variations through explicit temporal comparison. These results suggest that current MLLMs struggle to reliably distinguish fine-grained temporal changes across many urban indicators.



\textbf{Trend predictability varies substantially across indicators.} While several indicators remain close to chance-level performance, others are considerably more predictable. LUD achieves the highest accuracy across multiple models, with several models exceeding 0.75 accuracy. Urban form indicators such as RL and RD also achieve relatively strong performance. In contrast, many environmental and subjective perception-related indicators remain clustered around moderate accuracy levels. These patterns suggest that indicators with more direct visual manifestations are more predictable from temporal image sequences, whereas subtle environmental and perceptual changes remain difficult to identify reliably.

\textbf{Model scale provides limited advantage in temporal trend reasoning.} Performance differences across model families and parameter scales are inconsistent. Larger variants do not systematically outperform smaller counterparts within the same family, and improvements on one indicator are often offset by declines on others. This contrasts with the partial scaling benefits observed in single-year estimation. The relatively homogeneous performance across models suggests that temporal trend detection remains challenging regardless of model scale, highlighting a common limitation in capturing subtle temporal changes from image sequences.


To better understand these failures, we inspect model predictions and observe a consistent tendency toward temporal over-smoothing. When year-to-year differences are subtle, MLLMs frequently predict \textit{stable} even when the ground-truth trend is positive or negative. This behavior suggests that current MLLMs rely primarily on static visual cues and struggle to perform explicit temporal comparison across image sequences.

\subsection{Detailed Analysis}

In this section, we analyze performance variation across cities and investigate how different visual, temporal, and adaptation factors affect model performance. We further compare model performance against human baselines.

\begin{figure}[htb]
  \centering
    \subfigure[]{
    \centering
    \includegraphics[width=\linewidth]{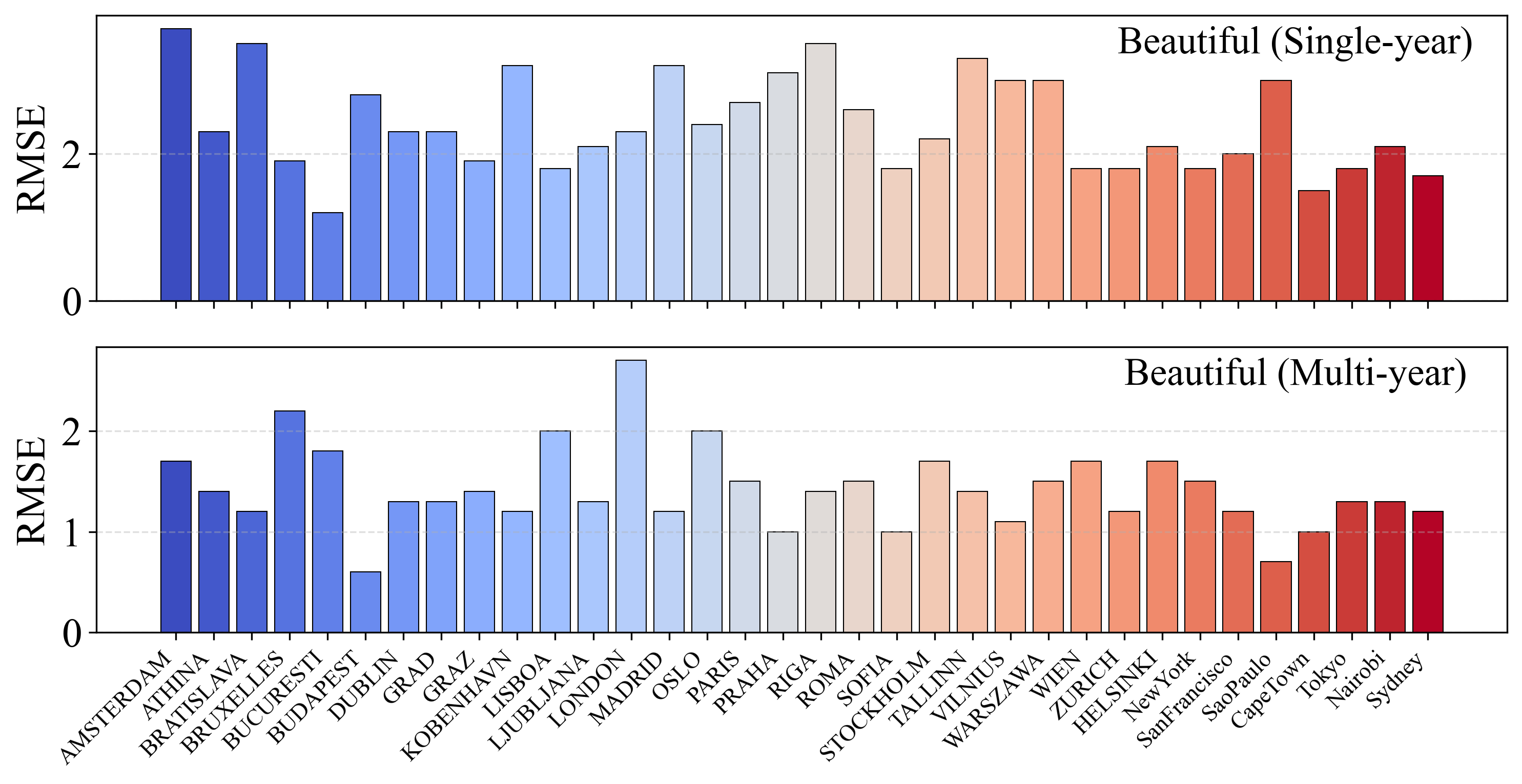}
    \label{ablation-city}
}

  \subfigure[]{
    \centering
    \includegraphics[width=0.475\linewidth]{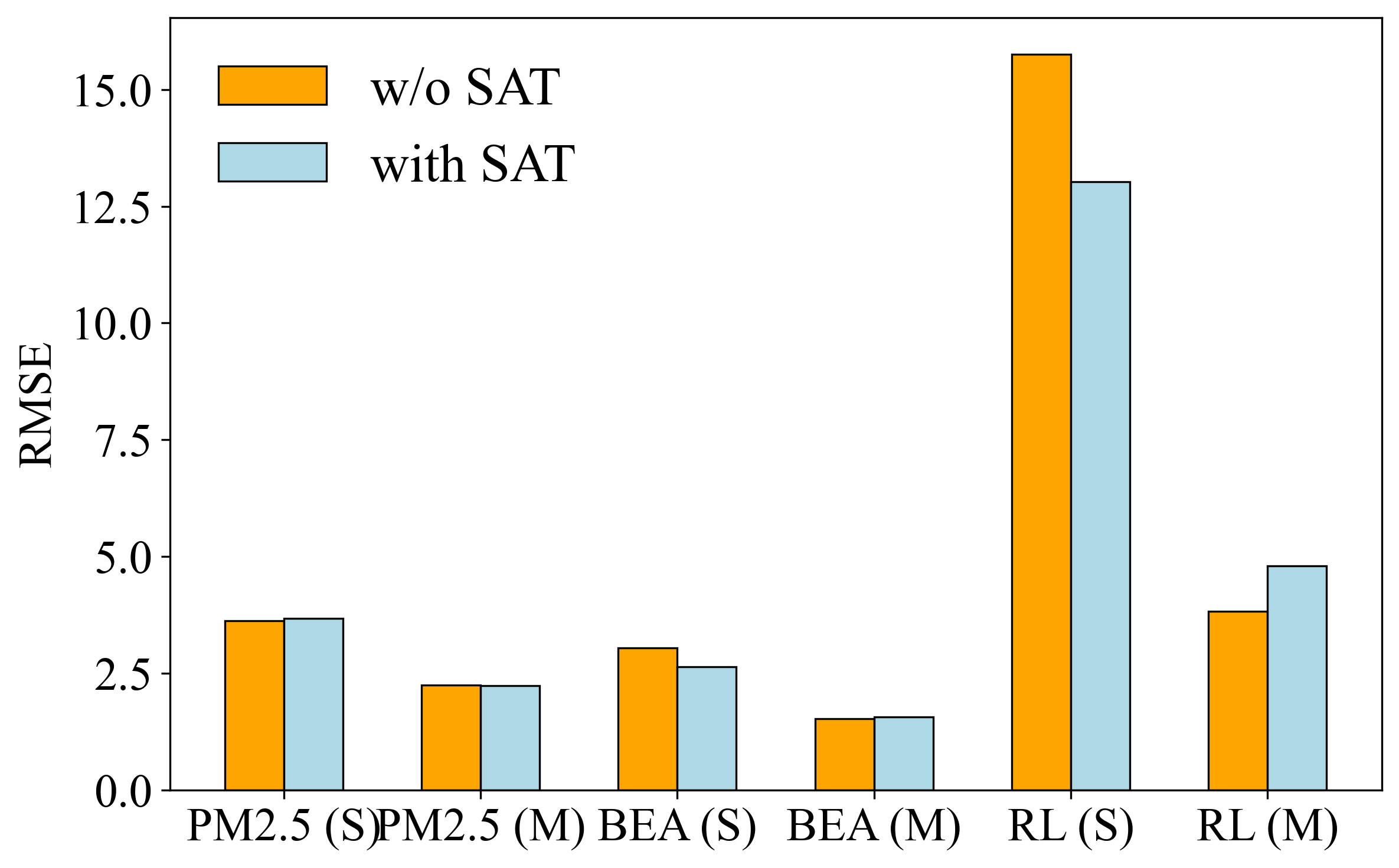}
    \label{ablation-sat}
}
    \subfigure[]{
    \centering
    \includegraphics[width=0.475\linewidth]{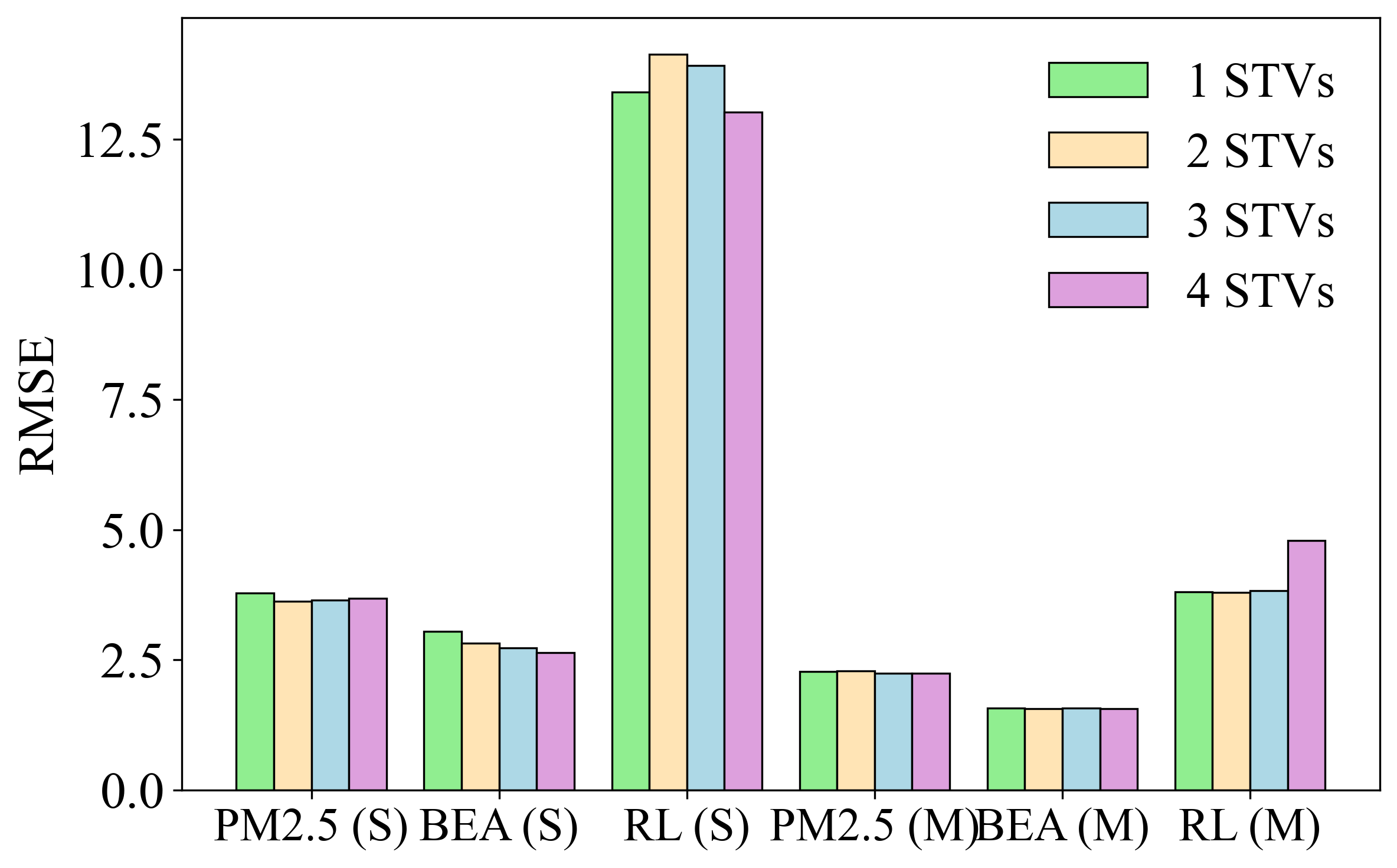}
    \label{ablation-stv}
}
    \subfigure[]{
    \centering
    \includegraphics[width=0.475\linewidth]{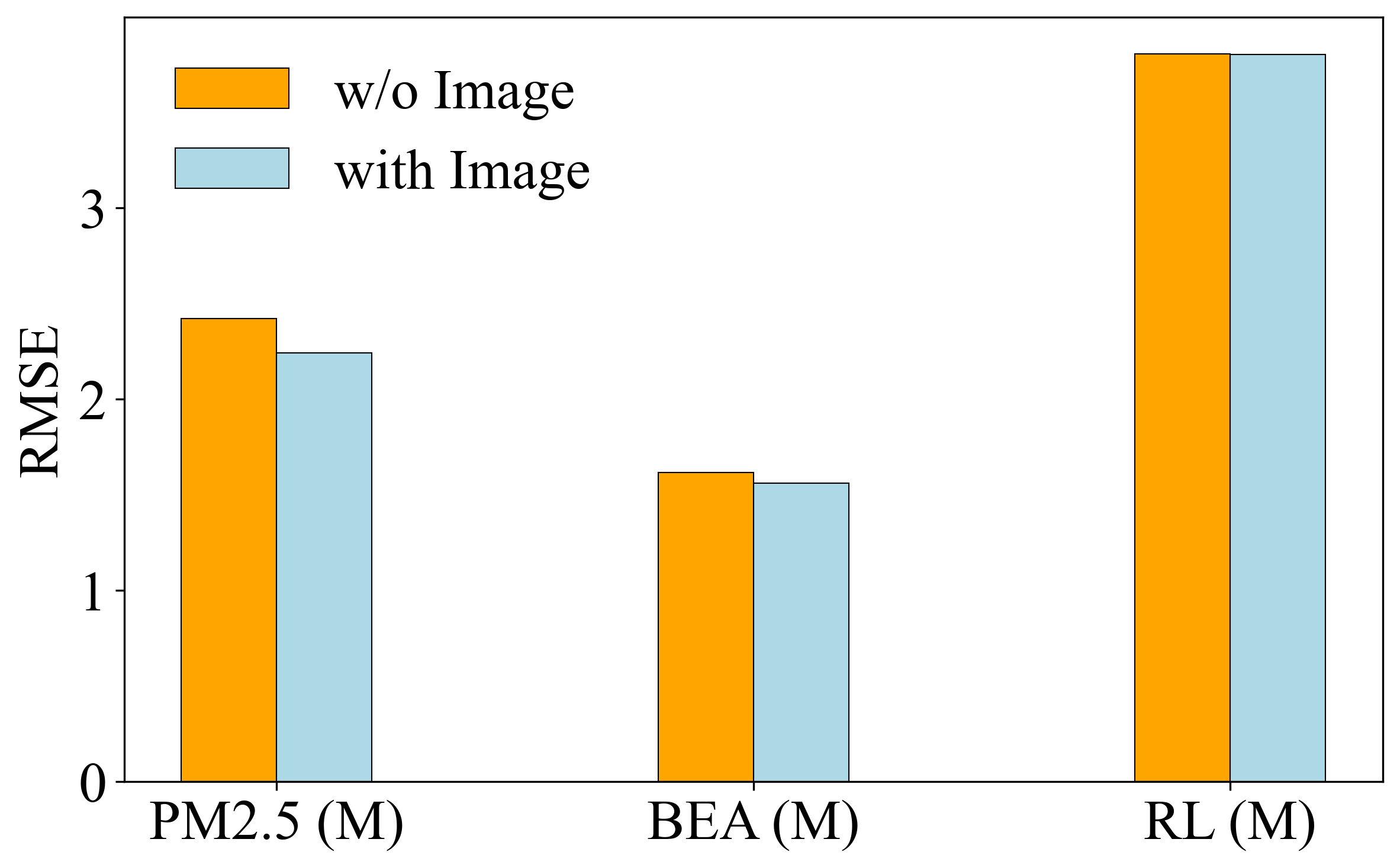}
    \label{ablation-modality}
}
    \subfigure[]{
    \centering
    \includegraphics[width=0.475\linewidth]{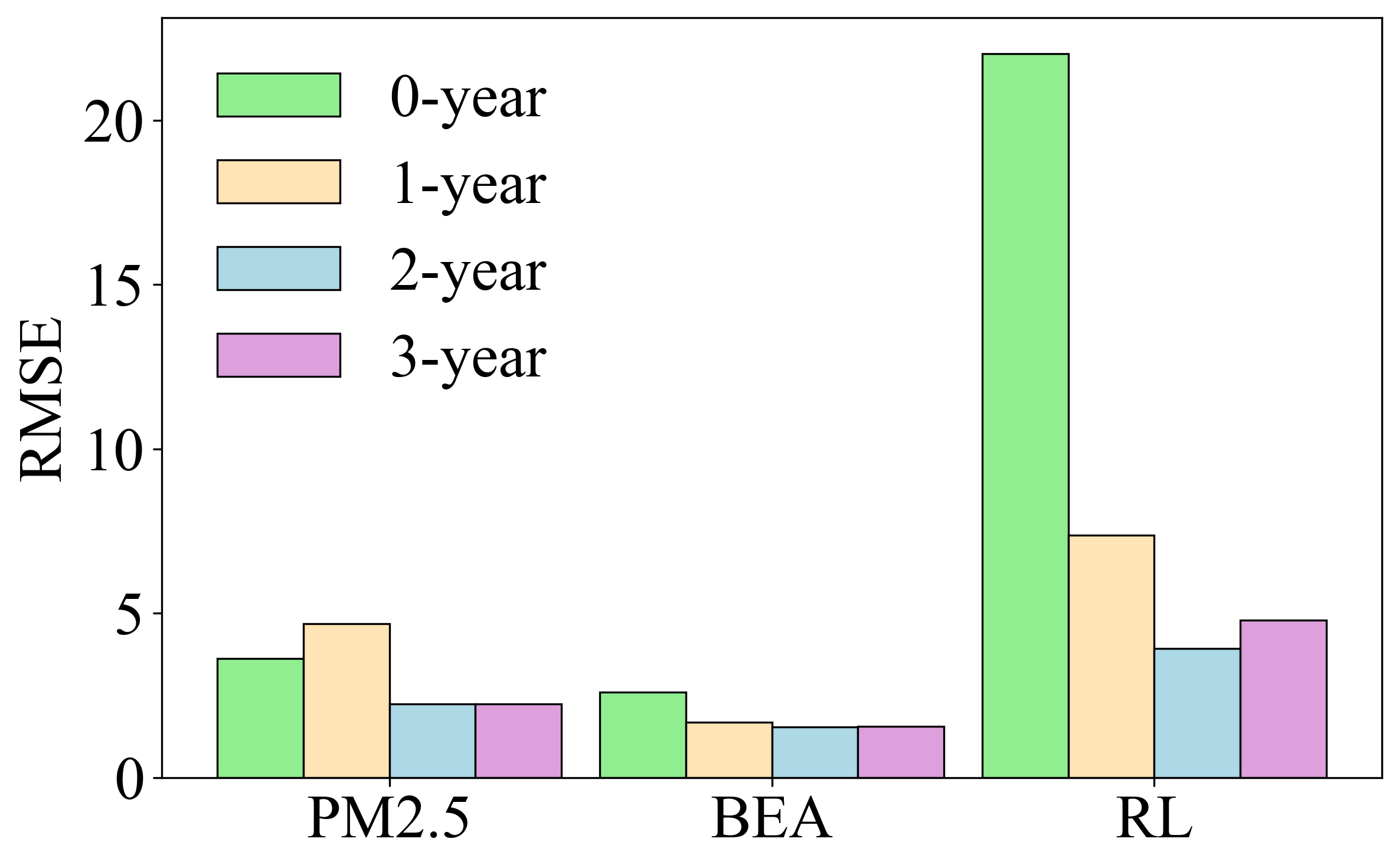}
    \label{ablation-year}
}
  \caption{RMSE results for (a) city-level \textit{beautiful} estimation and analyses of the effects of (b) satellite imagery, (c) street view image quantity, (d) image modality, and (e) temporal length. `S’ denotes the single-year estimation task, and `M’ denotes the multi-year forecasting task.}
  \vspace{-20px}
  \label{ablation-sat-stv-year}
\end{figure}

\textbf{City-Level Performance Variations.} 
Using Gemma3-27B as a representative model, we analyze city-level prediction error for the BEA indicator (Figure~\ref{ablation-city}). RMSE varies noticeably across cities in both single-year and multi-year settings, indicating heterogeneous prediction difficulty under different urban environments. Compared with single-year estimation, multi-year forecasting generally produces lower RMSE with reduced cross-city variance, suggesting that temporal context improves prediction stability across urban environments. Nevertheless, certain cities remain consistently challenging across both settings, implying that city-specific visual characteristics and urban heterogeneity continue to affect model performance even when temporal information is available.

\textbf{Effect of Satellite Imagery.}
We evaluate the contribution of satellite imagery by comparing two configurations: (i) satellite + street view imagery (with SAT) and (ii) street view imagery only (w/o SAT). Figure~\ref{ablation-sat} reports results for PM${2.5}$, BEA, and RL using Gemma3-27B. The effect of satellite imagery is indicator- and task-dependent. Environmental indicators exhibit limited sensitivity to aerial observations, whereas perceptual and structural indicators generally benefit more from additional spatial context. The gains are particularly evident in the single-year setting, while the contribution of satellite imagery becomes less consistent for multi-year forecasting. Overall, these results suggest that aerial context is most useful when the target indicator is visually grounded and the task relies on static spatial structure.


\textbf{Effect of Street View Image Quantity.}
We analyze the impact of the number of street view images per spatial unit (1--4 images) on PM${2.5}$, BEA, and RL (Figure~\ref{ablation-stv}). Increasing the number of street view images does not consistently improve performance across indicators. Environmental indicators remain largely insensitive to additional street-level coverage, while perceptual attributes benefit from more diverse visual perspectives. Structural indicators exhibit non-monotonic behavior, where adding more views may improve performance in some cases but also introduce redundancy or noisy visual signals. These results suggest that additional street view observations provide complementary information to varying degrees across different indicator types.


\textbf{Effect of Image Modality.}
We assess the contribution of visual modalities by comparing multi-year forecasting performance with and without image inputs (Figure~\ref{ablation-modality}). Incorporating satellite and street view imagery improves performance for certain indicators, particularly PM${2.5}$ and BEA, while yielding marginal changes for RL. These results suggest that visual observations provide complementary information beyond historical numerical inputs, but their contribution varies across indicators.

\textbf{Effect of Temporal Length.}
We examine the effect of the number of historical years provided as input (Figure~\ref{ablation-year}). Across all three indicators, incorporating temporal context generally reduces RMSE compared to using no historical input, confirming the importance of historical observations for multi-year forecasting. However, the benefit of longer temporal sequences is not monotonic. PM${2.5}$ and RL exhibit fluctuating performance as more years are included, whereas the BEA indicator improves consistently and then stabilizes. These results suggest that temporal context provides strong predictive signals, but longer sequences do not uniformly translate into better forecasting performance.

\textbf{Effect of Fine-Tuning.}
To examine whether the zero-shot setting underestimates model capability, we perform parameter-efficient LoRA fine-tuning \cite{hu2022lora} on Qwen2.5VL-7B using UrbanWell samples excluded from the benchmark split. Specifically, we keep the pretrained backbone frozen and train only the LoRA adapter parameters, then load the learned adapter on top of the same base model for multimodal inference. Table~\ref{tab:lora} reports results on five representative indicators covering each benchmark domain. Fine-tuning substantially improves performance in the single-year setting, particularly for Popu and CO$_2$ estimation, while gains in the multi-year forecasting setting are comparatively smaller. These results suggest that adaptation improves absolute performance but does not eliminate the heterogeneous difficulty across indicators, implying that the observed challenges extend beyond the zero-shot setting.

\begin{table}[htb]
\centering
\small
\setlength{\tabcolsep}{4pt}
\caption{Performance comparison between zero-shot and LoRA fine-tuned Qwen2.5VL-7B across representative UrbanWell indicators. Lower RMSE indicates better performance. `S' and `M' denote single-year estimation and multi-year forecasting, respectively.}
\label{tab:lora}
{
\begin{tabular}{l l c c c c c}
\toprule
& Model & Popu & RD & CO$_2$ & SPM & BEA \\
\midrule
S & Zero-shot & 7.20e4 & 16.43 & 1.10e8 & 8.07 & 3.35 \\
S & LoRA& \textbf{7942} & \textbf{13.35} & \textbf{1898} & \textbf{2.13} & \textbf{2.13} \\
\midrule
M & Zero-shot & 370 & 4.81 & \textbf{14.20} & 0.30 & 1.42 \\
M & LoRA& \textbf{319} & \textbf{4.04} & 15.12 & \textbf{0.27} & \textbf{1.35} \\
\bottomrule
\end{tabular}
}
\end{table}


\textbf{Comparison with Human Performance.}
We additionally evaluate human participants on subsets of four single-year estimation tasks (CO$_2$, Popu, BEA, and WEA). Humans participants achieve lower RMSE than the best-performing MLLM on CO$_2$ (50 vs.\ 175) and Popu (2901 vs.\ 6400), while remaining competitive on BEA (2.38 vs.\ 2.04) and WEA (1.84 vs.\ 1.69). These results reveal varying gaps between human and MLLM performance across different urban indicators.




Additional analyses on validation of pretrained subjective perception labels, prompt sensitivity, temporal forecasting baselines, and failure cases are provided in Appendices~\ref{subjective-labels}--\ref{failure-cases}.

\section{Related Work}

\subsection{MLLMs in Urban Applications}

Recent advances in MLLMs have enabled joint modeling of visual and textual information, supporting cross-modal reasoning over complex inputs~\cite{zhang2024mm}. Representative models such as Qwen \cite{bai2025qwen2}, GPT \cite{OpenAIGPT412025}, and InternVL \cite{zhu2025internvl3} demonstrate strong performance in image understanding \cite{bian2025icg}, cross-modal alignment \cite{ye2024loki}, and visual reasoning tasks \cite{zhang2025vrest}. While these models show robust generalization across general-purpose benchmarks \cite{li2024llava}, systematic evaluation in structured urban settings remains limited.

In urban applications, MLLMs have been explored for urban perception, spatial reasoning, and multimodal urban understanding~\cite{li2024can,feng2025citygpt,chen2025spatialllm,zhang2025urban}. UrbanCLIP \cite{yan2024urbanclip} aligns satellite imagery with textual descriptions using contrastive learning, and UrbanVLP~\cite{hao2025urbanvlp} leverages the descriptive capabilities of MLLMs to bridge satellite and street view imagery for urban prediction. UrbanLLaVA \cite{feng2025urbanllava} further integrates satellite and street view imagery with structured urban datasets for improved generalization on urban tasks. However, existing efforts primarily focus on static prediction settings. Standardized benchmarks for evaluating temporal reasoning over longitudinal urban dynamics in MLLMs remain limited.

\subsection{Urban Visual Dataset and Benchmark}

Urban imagery, such as satellite and street view imagery, has been leveraged to quantify spatial patterns, urban dynamics, and socioeconomic conditions~\cite{liu2023knowledge,li2022predicting}. Satellite imagery captures large-scale spatial structure from a bird’s-eye perspective~\cite{xi2022beyond,yan2024urbanclip}, while street view imagery complements satellite observations with fine-grained ground-level visual information \cite{fan2023urban,2024_global_streetscapes}. Building upon these multimodal urban imagery, recent benchmarks have begun to evaluate the capabilities of MLLMs in urban understanding and reasoning tasks. CityBench \cite{feng2025citybench} focuses on urban navigation and identity tasks, UrBench \cite{zhou2025urbench} integrates multi-view imagery for spatial reasoning, CityLens \cite{liu2025citylens} incorporates socioeconomic data for urban function modeling and UrbanFeel \cite{he2025urbanfeel} introduces temporal sequences with perceptual indicators. CityEQA \cite{zhao2025cityeqa} studies embodied urban question answering, while MapDR \cite{chang2025driving} aligns traffic semantics with structured map representations.

Despite these advances, existing benchmarks primarily focus on perception, navigation, or task-specific urban reasoning, with limited support for comprehensive urban wellbeing analytics. In particular, few benchmarks jointly model multi-year satellite and street view imagery together with heterogeneous indicators under a unified evaluation framework. Moreover, temporal urban reasoning remains underexplored, especially for forecasting future urban conditions and understanding longitudinal urban trends across cities. UrbanWell addresses these limitations through a large-scale multi-year benchmark that systematically evaluates spatio-temporal urban wellbeing reasoning from satellite and street view imagery.

\section{Ethical Use of Data}

UrbanWell is constructed from satellite imagery, street view imagery, and publicly available geospatial datasets. The spatial resolution of the satellite imagery is insufficient for identifying individuals, while street view images are obtained from officially released platforms where sensitive information has been anonymized prior to publication. All derived indicators are aggregated at grid level and contain no personally identifiable information. Consistent with prior urban computing research using similar data sources \cite{liu2025citylens,he2025urbanfeel}, we comply with applicable platform policies and data usage regulations. To respect intellectual property and licensing constraints, UrbanWell does not redistribute raw satellite or street view imagery. Instead, we release benchmark annotations, geographic metadata, timestamps, and a deterministic retrieval pipeline enabling reconstruction of the benchmark inputs from the original providers under their respective terms of service. This design supports reproducible evaluation and standardized comparison across future methods.

\section{Conclusion}

We introduce UrbanWell, a large-scale multimodal benchmark for evaluating spatio-temporal reasoning capabilities of MLLMs in urban wellbeing analytics. UrbanWell spans 38 cities and 19 heterogeneous urban indicators, integrating multi-year satellite and street view imagery with grid-level socio-environmental measurements. 
The benchmark provides standardized evaluation framework for single-year estimation, multi-year forecasting, and temporal trend classification. Extensive experiments on 15 state-of-the-art MLLMs reveal substantial performance variation across indicators, task settings, and model architectures. The results further show that, although current models perform relatively well on general-purpose visually grounded tasks, they remain challenged by environmental estimation, demographic inference, and fine-grained temporal reasoning, highlighting limitations in multimodal quantitative and longitudinal urban understanding. These findings underscore the need for more robust multimodal reasoning capabilities in urban-scale AI systems and motivate future research on multimodal spatio-temporal reasoning for urban analytics.



\textbf{Limitations \& Future Work.} Subjective perception indicators in UrbanWell are derived from pretrained vision models trained on human-labeled datasets~\cite{dubey2016deep}, and may therefore inherit biases present in the original annotations. Future extensions may incorporate additional human-validated signals and cross-cultural calibration to improve robustness of perception-related evaluation. UrbanWell currently covers 38 cities, primarily in Europe, and future work will extend the benchmark to more geographically diverse and underrepresented urban regions with richer temporal observations and broader multimodal coverage. A further limitation concerns temporal alignment, as imagery captures specific moments while some indicators are annual aggregates. Although year-level alignment is common in urban computing and urban environments are generally stable at this timescale, some mismatch may still exist. Finally, UrbanWell evaluates MLLMs as deployed, including knowledge potentially acquired during pretraining. While the benchmark is constructed from grid-level imagery and year-specific indicators, making direct memorization unlikely, potential exposure to related geographic information during pretraining cannot be completely excluded. Future work will also investigate more robust multimodal temporal reasoning and calibration strategies for urban-scale quantitative prediction.

\section{Acknowledgments}

This work was supported in part by NordForsk through Nordic University Cooperation on Edge Intelligence (168043), in part by Finnish Foundation for Technology Promotion Grant (11246), in part by City of Helsinki Research Grant (2025), in part by Research Council of Finland XRISE project (371849), and in part by the Zhongguancun Academy (Grant No. XTS0074).

\bibliographystyle{ACM-Reference-Format}
\bibliography{sample-base}

\appendix
\section{Appendix}
\subsection{List of Covered Cities} \label{app_city_names}
UrbanWell covers 38 cities: Amsterdam, Athens, Belgrade, Berlin, Bratislava, Brussels, Bucharest, Budapest, Cape Town, Copenhagen, Dublin, Graz, Helsinki, Lisbon, Ljubljana, London, Madrid, Nairobi, New York, Oslo, Paris, Prague, Riga, Rome, San Francisco, Sao Paulo, Skopje, Sofia, Stockholm, Sydney, Tallinn, Tirana, Tokyo, Vienna, Vilnius, Warsaw, Zagreb, Zurich.

\subsection{Indicator Formulation} \label{data-processing-appendix}
\textbf{Land Use Diversity (LUD).} The land use diversity indicator for grid $r$ is computed as~\cite{scepanovic2021jane}:
\begin{equation}
\text{LUD}_r = - \frac{\sum_{j=1}^{n_{LUC}} P_{r,j} \log(P_{r,j})}{\log(n_{LUC})},
\end{equation}
where $P_{r,j}$ denotes the proportion of land use category $j$ within grid $r$, and $n_{LUC}=3$ is the total number of land use categories: (i) residential areas, (ii) commercial, industrial, institutional, and governmental areas, and (iii) recreational spaces, parks, and water bodies.

\textbf{Economic Activity Diversity (EcD).} The economic activity diversity indicator for grid $r$ is computed using Shannon entropy~\cite{shannon1948mathematical}:
\begin{equation}
\text{EcD}_r = - \sum_{k=1}^{n_{POI}} P_{r,k} \log(P_{r,k}),
\end{equation}
where $P_{r,k}$ denotes the proportion of POIs belonging to category $k$ within grid $r$, and $n_{POI}$ is the total number of commercial POI categories. The commercial POI categories include restaurants, fast food outlets, cafes, bars, pubs, supermarkets, convenience stores, malls, clothing stores, shoe stores, electronics stores, jewellers, bakeries, butcher shops, florists, bookshops, hairdressers, beauty shops, laundries, repair shops, photo studios, travel agencies, car rentals, car washes, hotels, motels, guesthouses, hostels, campsites, cinemas, theatres, museums, galleries, nightclubs, casinos, sports centres, and fitness centres.


\begin{table*}[htb]
  \centering
  \caption{Prompt templates used for single-year estimation, multi-year forecasting, and temporal trend analysis in UrbanWell.}
  \label{tab-prompt}
  \begin{tabularx}{\textwidth}{p{0.18\textwidth} X}
    \toprule
    Task Paradigm & Prompt Template \\
    \midrule

    Single-year Estimation &
    Suppose you are a vision expert specializing in socioeconomic and environmental analysis. Examine \texttt{\{number\_of\_total\_images\}} images from the same region: 1 satellite image from year \texttt{\{year\}}, and \texttt{\{number\_of\_street\_view\_images\}} street views from the same year. Analyze the imagery content, and provide your assessment of \texttt{\{indicator\}} in \texttt{\{indicator\_unit\}} according to the \texttt{\{measurement\_definition\}} provided by its data source. Reply only with the numeric value. Nothing else should be included. \\
    \midrule

    Multi-year Forecasting &
    Suppose you are a vision expert. You are given a series of images for a region spanning \texttt{\{number\_of\_total\_years\}} years, from \texttt{\{first\_year\}} to \texttt{\{last\_year\}}. For each historical year, you are provided with 1 satellite image, \texttt{\{number\_of\_street\_view\_images\}} street view image(s), and the corresponding \texttt{\{indicator\}} in \texttt{\{indicator\_unit\}} based on the \texttt{\{measurement\_definition\}}. For the target year \texttt{\{target\_year\}}, You are provided with only the images. Using the historical years as context, estimate \texttt{\{indicator\}} in \texttt{\{indicator\_unit\}} for \texttt{\{target\_year\}}. Reply only with the numeric value. \\
    \midrule

    Temporal Trend Analysis &
    You are given images for a region across \texttt{\{number\_of\_total\_years\}} years from \texttt{\{first\_year\}} to \texttt{\{last\_year\}} (each year includes 1 satellite image and \texttt{\{number\_of\_street\_view\_images\}} street view image(s)). For each consecutive two-year step, classify the trend of \texttt{\{indicator\}} in \texttt{\{indicator\_unit\}} according to the \texttt{\{measurement\_definition\}} as \texttt{positive}, \texttt{negative}, or \texttt{stable} based on predefined relative-change thresholds. Reply only with the label. \\
    \bottomrule
  \end{tabularx}
\end{table*}



\subsection{Task Prompt Templates} \label{prompt-example}
Table~\ref{tab-prompt} summarizes the prompt templates used in UrbanWell for the three task paradigms: Single-year Estimation, Multi-year Forecasting, and Temporal Trend Analysis. Each task adopts a standardized prompt structure specifying the input imagery, temporal context, historical observations, and expected output format.

For single-year estimation, the prompt provides satellite and street view imagery from a specific year and asks the model to estimate a target urban indicator. For multi-year forecasting, the prompt includes sequential imagery together with historical indicator values to predict the target value in the final year. For temporal trend analysis, the prompt presents imagery from consecutive years and requires classification of the indicator trend according to predefined thresholds. All prompts follow unified templates with fixed placeholders to ensure controlled and reproducible evaluation across tasks and models. Additional prompt examples and corresponding MLLM outputs for all task paradigms are provided in the project repository: \url{https://github.com/axin1301/UrbanWell-Benchmark}.





\subsection{Validation of Subjective Perception Labels} \label{subjective-labels}
Since subjective perception indicators are derived from pretrained models rather than direct human annotations, we compare benchmark labels against a subset of 100 human-annotated samples. For the Qwen-series models, evaluation against human annotations and pretrained scores yields highly comparable average RMSE (BEA: 4.3 vs.\ 4.7; WEA: 4.8 vs.\ 4.9), suggesting that the proxy labels remain closely aligned with human perception for large-scale benchmarking.


\subsection{Comparison with Temporal Baselines} \label{temporal-baselines}

To contextualize MLLM performance, we additionally evaluate several temporal forecasting baselines, including ARIMA, Ridge Regression (RR), LSTM \cite{hochreiter1997long}, PatchTST \cite{nie2022time}, and MICN \cite{wang2023micn}, using the same historical indicator observations. As shown in Table~\ref{tab:forecast_baseline}, temporal baselines remain competitive for population forecasting, while MLLMs achieve lower RMSE on CO$_2$ and RD prediction. For the perception-related indicator (BEA), Ridge Regression achieves the best performance. These results suggest that temporal models remain effective for certain urban indicators.

\begin{table}[htb]
\centering
\caption{Comparison between MLLMs and temporal forecasting baselines on representative indicators. Lower RMSE indicates better performance.}
\label{tab:forecast_baseline}
\small
\setlength{\tabcolsep}{4pt}
\begin{tabular}{lcccc}
\toprule
Model & Popu & CO$_2$ & BEA & RD \\
\midrule
ARIMA & \textbf{179} & 16 & 1.92 & 3.65 \\
RR & 5826 & 130 & \textbf{1.28} & 8.85 \\
LSTM & 5301 & 127 & 1.36 & 8.66 \\
PatchTST & 6872 & 140 & 1.80 & 9.79 \\
MICN & 5320 & 131 & 1.31 & 8.34 \\
\midrule
Best MLLM & 200 & \textbf{11.5} & 1.35 & \textbf{2.56} \\
\bottomrule
\end{tabular}
\end{table}


\subsection{Prompt Sensitivity Analysis} \label{prompt-variations}

To assess prompt sensitivity, we evaluate two alternative prompt variants: (1) a reasoning-enhanced prompt that encourages step-by-step inference, and (2) a structured prompt with explicit formatting and concise instructions. Table~\ref{tab:prompt} reports results for Gemma3-27B on representative indicators under both single-year estimation and multi-year forecasting settings. Structured prompts generally improve performance for certain indicators, such as CO$_2$ and Popu in the single-year setting, likely because the default prompts intentionally provide minimal task guidance. However, the gains are inconsistent across tasks and indicators. Reasoning-enhanced prompts often substantially degrade performance, particularly for multi-year forecasting tasks.

\begin{table}[htb]
\centering
\caption{Prompt sensitivity analysis across representative indicators under single-year estimation (S) and multi-year forecasting (M). Lower RMSE indicates better performance.}
\label{tab:prompt}
\small
\setlength{\tabcolsep}{4pt}
\begin{tabular}{l c c c c}
\toprule
Indicator & Setting & Default & Reasoning & Structured \\
\midrule
CO$_2$ & S & 371 & 1531 & \textbf{261} \\
Popu & S & 16312 & 12792 & \textbf{5031} \\
BEA & S & \textbf{2.7} & 4.0 & 4.1 \\
\midrule
CO$_2$ & M & \textbf{10} & 596 & 11 \\
Popu & M & 229 & 9965 & \textbf{179} \\
BEA & M & 1.5 & 1.5 & 1.5 \\
\bottomrule
\end{tabular}
\end{table}

\subsection{Failure Case Analysis} \label{failure-cases}

We observe several recurring failure patterns across tasks. Indicators that are not directly observable from imagery, such as CO$_2$ and Popu, often exhibit large estimation errors in the single-year setting due to the lack of strong visual signals. For example, a grid with ground-truth CO$_2$ value 89 may be predicted as 12. In temporal tasks, models frequently fail to capture subtle urban changes, such as gradual population-related densification, and often predict \textit{stable} when year-to-year differences are visually minor.

\end{document}